\documentclass[11pt,a4paper]{article}
\usepackage[hyperref]{ms}
\usepackage{times}
\usepackage{latexsym}

\usepackage{url}
\usepackage{amsmath}
\usepackage{tabularx}
\usepackage{siunitx}
\usepackage{amsfonts}  
\usepackage{graphicx}
\usepackage[inline]{enumitem}
\usepackage{array,colortbl,multirow,multicol,booktabs,ctable}
\usepackage{breqn}

\aclfinalcopy 

\title{Spherical Latent Spaces for Stable Variational Autoencoders}

\author{Jiacheng Xu ~{\normalfont and}~ Greg Durrett \\
 Department of Computer Science \\
 The University of Texas at Austin\\
 {\tt \{jcxu,gdurrett\}@cs.utexas.edu}}

\date{}

\begin{document}
\maketitle

\begin{abstract}
A hallmark of variational autoencoders (VAEs) for text processing is their combination of powerful encoder-decoder models, such as LSTMs, with simple latent distributions, typically multivariate Gaussians. These models pose a difficult optimization problem: there is an especially bad local optimum where the variational posterior always equals the prior and the model does not use the latent variable at all, a kind of ``collapse'' which is encouraged by the KL divergence term of the objective. In this work, we experiment with another choice of latent distribution, namely the von Mises-Fisher (vMF) distribution, which places mass on the surface of the unit hypersphere. With this choice of prior and posterior, the KL divergence term now only depends on the variance of the vMF distribution, giving us the ability to treat it as a fixed hyperparameter. We show that doing so not only averts the KL collapse, but consistently gives better likelihoods than Gaussians across a range of modeling conditions, including recurrent language modeling and bag-of-words document modeling. An analysis of the properties of our vMF representations shows that they learn richer and more nuanced structures in their latent representations than their Gaussian counterparts.\footnote{The code and dataset are available at: \url{https://github.com/jiacheng-xu/vmf_vae_nlp}}

\end{abstract}

\section{Introduction}

Recent work has established the effectiveness of deep generative models for a range of tasks in NLP, including text generation \cite{hu2017toward,yu2017seqgan}, machine translation \cite{zhang2016variational}, and style transfer \cite{shen2017style,junbo2017adversarially}.
Variational autoencoders, which have been explored in past work for text modeling \cite{miao2016neural,bowman2016generating}, posit a continuous latent variable which is used to capture latent structure in the data.
Typical VAE implementations assume the prior of this latent space is a multivariate Gaussian; during training, a Kullback-Leibler (KL) divergence term in loss function encourages the variational posterior to approximate the prior. One major limitation of this approach observed by past work is that the KL term may encourage the posterior distribution of the latent variable to ``collapse'' to the prior, effectively rendering the latent structure unused \cite{bowman2016generating,chen2016variational}. 

In this paper, we propose to use the von Mises-Fisher (vMF) distribution rather than Gaussian for our latent variable.  vMF places a distribution over the unit hypersphere governed by a mean parameter $\mu$ and a concentration parameter $\kappa$. Our prior is a uniform distribution over the unit hypersphere ($\kappa = 0$) and our family of posterior distributions treats $\kappa$ as a \emph{fixed} model hyperparameter. Since the KL divergence only depends on $\kappa$, we can structurally prevent the KL collapse and make our model's optimization problem easier. We show that this approach is actually more robust than trying to flexibly learn $\kappa$, and a wide range of settings for fixed $\kappa$ lead to good performance. Our model systematically achieves better log likelihoods than analogous Gaussian models while having higher KL divergence values, showing that it more successfully makes use of the latent variables at the end of training.

\begin{figure*}[t!]
\centering
\includegraphics[width=1\textwidth]{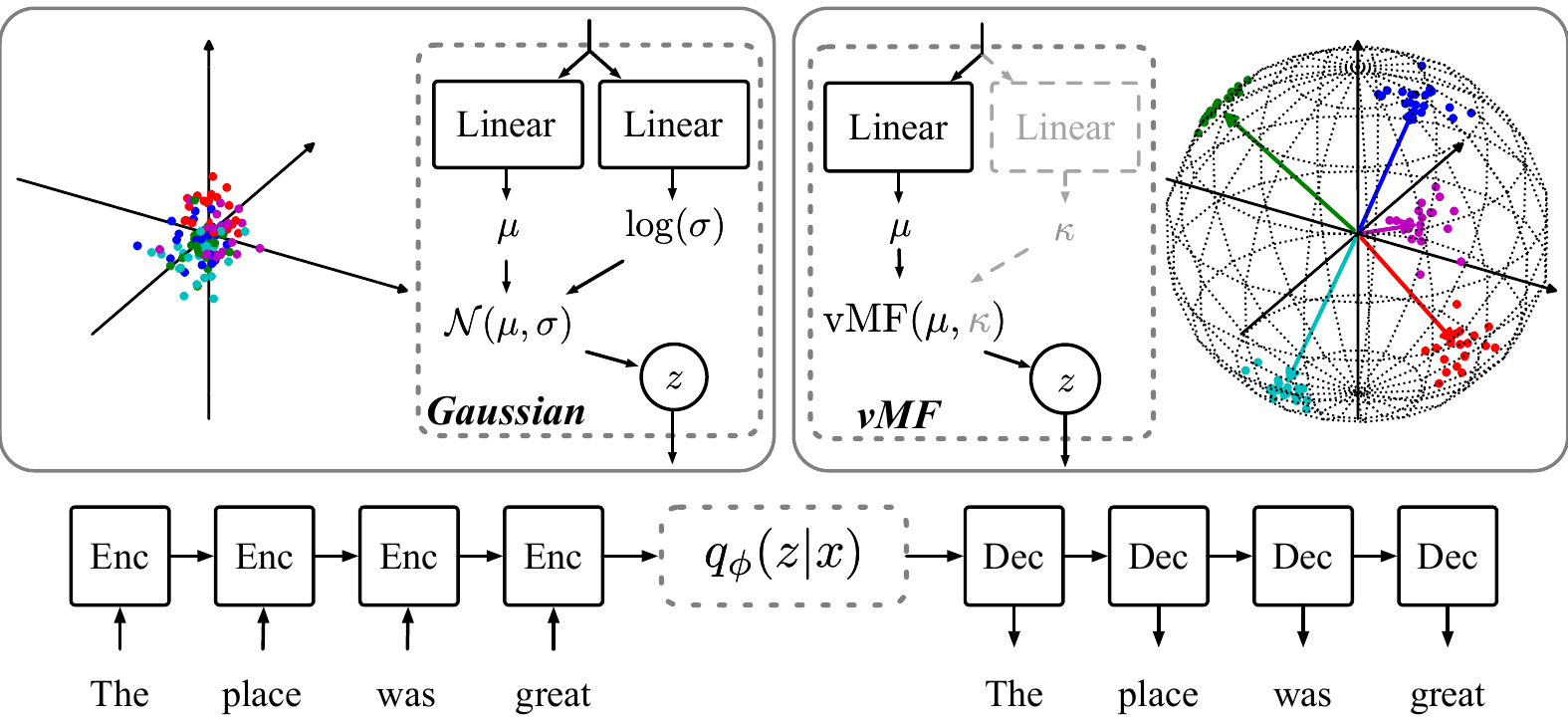}
\caption{The Neural Variational RNN (NVRNN) language model based on a Gaussian prior (left) and a vMF prior (right). The encoder model first computes the parameters for the variational approximation $q_\phi(z|x)$ (see dotted box); we then sample $z$ and  generate the word sequence $x$ given $z$. We show samples from $\mathcal{N}(0,I)$ and vMF($\cdot$,$\kappa=100$); the latter samples lie on the surface of the unit sphere. While $\kappa$ can be predicted from the encoder network, we find experimentally that fixing it leads to more stable optimization and better performance.}
\label{model}
\end{figure*}

Past work has suggested several other techniques for dealing with the KL collapse in the Gaussian case. Annealing the weight of KL term \cite{bowman2016generating} still leaves us with brittleness in the optimization process, as we show in Section~\ref{sec:bg}. Other prior work \cite{yang2017improved,semeniuta2017a} focuses on using CNNs rather than RNNs as the decoder in order to weaken the model and encourage the use of the latent code, but the gains are limited and changing the decoder in this way requires ad hoc model engineering and careful tuning of various decoder capacity parameters. Our method is orthogonal to the choice of the decoder and can be combined with any of these approaches. Using vMF distributions in VAEs also leaves us the flexibility to modify the prior in other ways, such as using a product distribution with a uniform \cite{guu2017generating} or piecewise constant term \cite{serban2017piecewise}.



We evaluate our approach in two generative modeling paradigms. For both RNN language modeling and bag-of-words document modeling, we find that vMF is more robust than a Gaussian prior, and our model learns to rely more on the latent variable while achieving better held-out data likelihoods. To better understand the contrast between these models, we design and conduct a series of experiments to understand the properties of the Gaussian and vMF latent code spaces, which make different structural assumptions. Unsurprisingly, these latent code distributions capture much of the same information as in a bag of words, but we show that vMF can more readily go beyond this, capturing ordering information more effectively than a Gaussian code.

\section{Variational Autoencoders for Text}
\label{sec:bg}

\newcite{bowman2016generating} propose a variational autoencoder model for generative text modeling inspired by \newcite{kingma2013autoencoding}. Instead of modeling $p(x)$ directly as in vanilla language models, VAEs introduce a continuous latent variable $z$ and take the form $p(z)p(x|z)$. To train a VAE, we optimize the marginal likelihood $p(x)=\int p_{\theta}(z) p(x|z)dz$. 
The marginal log likelihood can be written as:
\begin{displaymath}
\log p_\theta (x) = \text{KL}(q_{\phi}(z|x)||p_{\theta}(z|x))   +  \mathcal{L}(\theta,\phi;x)\\
\end{displaymath}
\begin{multline} 
\mathcal{L}(\theta,\phi;x) =  - \text{KL}(q_{\phi}(z|x) || p_{\theta}(z) ) \\
+  \mathbb{E}_{q_{\phi}(z|x)}\log p_{\theta}(x|z) \label{elbo}
\end{multline}
$q_{\phi}(z|x)$, a variational approximation to the posterior $p(z|x)$, can be variously interpreted as a \textit{recognition model} or encoder, parameterized by a neural network to encode the sentence $x$ into a dense \textit{code} $z$.
$\mathcal{L}(\theta,\phi;x)$ is often called the evidence lower bound (ELBO). The first term of ELBO is the KL divergence of the approximate posterior from prior and the second term is an expected reconstruction error.

Since KL divergence is always non-negative, we can use $\mathcal{L}(\theta,\phi;x)$ as a lower bound of marginal likelihood $\log p_{\theta}(x)$. We optimize $\mathcal{L}(\theta,\phi;x)$, jointly learning the recognition model parameters $\phi$ and generative model parameters $\theta$.

As the choice of prior $p(z)$, most previous work uses a centered multivariate Gaussian $p_{\theta}(z)=\mathcal{N}(z;0,I)$. Since Gaussians are a location-scale family of distributions, using them for both the prior and posterior allows us to apply the reparameterization trick and differentiate through the sampling stage $z \sim \mathbb{E}_{q_\phi(z|x)}$ when optimizing ELBO in practice \cite{kingma2013autoencoding}.

\subsection{Case Study: NVRNN}
A Neural Variational RNN (NVRNN) for language modeling is described in \newcite{bowman2016generating} and depicted in Figure~\ref{model}. The goal of the NVRNN model is to extract a high level representation of a sentence into $z$ and reconstruct the sentence with a neural language model.

We denote a sequence of words as $x=\{x_1, x_2, \cdots,x_n\}$. 
Unlike in vanilla language modeling, an NVRNN conditions on the latent variable $z$ at each step of the generation $p_{\theta}(x|z) =  p_{\theta}(x_1|z) \prod_{i=1}^n p(x_i|x_1,\ldots,x_{i-1},z).$
This probability distribution is modeled using a recurrent model like an LSTM \cite{hochreiter1997long} as illustrated in Figure~\ref{model}. There is nothing unique about this choice; other recurrent sequence models like a CNN or a Transformer \cite{vaswani2017attention} could be used.


\subsection{Posterior Collapse}

When training a VAE, we update $\theta$ and $\phi$ simultaneously.  Optimizing Eq.~\ref{elbo} gives two gradient terms: an update from the reconstruction loss (likelihood of the correct labels) and an update from the KL divergence. While the reconstruction loss term encourages the $z$ to convey useful information to this model, the KL term consistently tries to regularize $q(z|x)$ towards the prior on every gradient update. This may trap the model in a bad local optimum where $q_{\phi}(z|x) = p_{\theta}(z)$ for all $x$: in this case, $z$ is simply a noise source, which is useless to the model, so the model has learned to ignore it and will not make large enough gradient updates to break $q(z|x)$ out of this optimum.


\newcite{bowman2016generating} termed this issue \textit{KL collapse} and proposed an annealing schedule to handle it, where the weight of the KL term is increased over the course of training.\footnote{Reweighting the KL term is also used in methods like $\beta$-VAE \cite{betavae} and InfoVAE \cite{Zhao_Song_Ermon_2017}.} In this way, the model initially learns to use the latent code but is then regularized towards the prior as training progresses. However, this trick is not sufficient to avert KL collapse in all scenarios, particularly when strong decoders are used and $z$ has a minor impact on $p_\theta(x|z)$.

Table~\ref{collapse} shows experiments in a similar setup to that of \newcite{bowman2016generating}. We train an NVRNN model on the Penn Treebank with four different hyperparameter settings. We either use a 3-layer LSTM encoder or a 1-layer LSTM and use or do not use a sigmoid annealing schedule (increase the KL weight from 0 to 1 over the first 20 epochs). We observe the best performance using the 1-layer model with annealing. 
One might conclude from this table that the annealing trick has worked since both models achieve better performance when annealing is used. But in fact, a vMF-based model can do better than either (NLL of 117), and moreover, we have no way of knowing that a better annealing scheme might not achieve even higher performance after training. Furthermore, the higher-capacity 3-layer model can theoretically do anything the 1-layer model can, so its lower performance indicates that our training is derailed either by overfitting or getting stuck in a local optimum where the latent variable is unused.\footnote{In our experiments, we found significant variance in collapse frequency due to other hyperparameters including whether the encoder is a unidirectional or bidirectional LSTM.}

\begin{table}[t]
\centering
\small
\begin{tabular}{@{}c|cc|cc@{}}
\toprule
       & \multicolumn{2}{c|}{No annealing} & \multicolumn{2}{c}{Sigmoid annealing} \\ 
      & 3-layer           & 1-layer          & 3-layer          & 1-layer          \\ \midrule
  KL  &            0.00 & 3.37 & 1.05 & \textbf{6.52}           \\ 
        NLL &     135	&129& 132	& \textbf{125}  \\ 
\bottomrule
\end{tabular}
\caption{Development set KL and NLL values for two NVRNN models trained on the Penn Treebank with and without the annealing technique of \newcite{bowman2016generating}. The higher-capacity 3-layer model collapses when no annealing is used, and while annealing works to improve performance, it still does not perform as well as the 1-layer variant. By contrast, a variant of the 1-layer model with vMF gives an NLL value of 117 and a KL of 18.6, a stronger result relying more heavily on the latent variable.}
\label{collapse}
\end{table}

Getting the best performance out of a VAE is, therefore, a challenging problem that requires careful tuning of the objective function and optimization procedure \cite{bowman2016generating,Zhao_Song_Ermon_2017,betavae}. Beyond the well-documented problem of KL collapse, an optimizer may simply get stuck in a local optimum during training and as a result, fail to find a model that most effectively exploits the latent variable.

The solution we advocate for in this paper is to change the distribution for the latent space and simplify the optimization problem. In the next section, we describe the von Mises-Fisher distribution and its use in VAE, where it forces the model to put the latent representations on the surface of the unit hypersphere rather than squeezing everything to the origin. Critically, this distribution lets us fix the value of the KL term by fixing the distribution's concentration parameter $\kappa$; this averts the KL collapse and leads to good model performance across two generative modeling paradigms.






\section{von Mises-Fisher VAE}
\label{sec:model}

The von Mises-Fisher distribution is a distribution on the $(d-1)$-dimensional sphere in $\mathbb{R}^d$. The vMF distribution is defined by a direction vector $\mu$ with $||\mu||=1$ and a concentration parameter $\kappa \geq 0$. The PDF of the vMF distribution for the $d$-dimensional unit vector $x$ is defined as:
\begin{align}
f_d (x;\mu, \kappa) = C_d(\kappa) \exp (\kappa \mu^{T} x)\\
C_d(\kappa) = \frac{ \kappa^{d/2-1} }{ (2\pi)^{d/2} I_{d/2-1}(\kappa)}
\end{align}
where $I_v$ stands for the modified Bessel function of the first kind at order $v$.

Figure~\ref{model} shows samples from vMF distributions with various $\mu$ vectors (arrows), $d=3$, and $\kappa=100$. This is a high $\kappa$ value, leading to samples that are tightly clustered around $\mu$, which is the mean and mode of the distribution. When $\kappa=0$, the distribution degenerates to a uniform distribution over the hypersphere independent of $\mu$.

Past work has used vMF as an emission distribution in unsupervised clustering models \cite{banerjee2005clustering}, VAE for other domains \cite{Davidson_Falorsi_Cao_Kipf_Tomczak_2018,Hasnat_Bohné_Milgram_Gentric_Chen_2017}, and a generative editing model for text \cite{guu2017generating}. We focus specifically on the empirical properties of vMF for text modeling and conduct a systematic examination of how this prior affects VAE models compared to using a Gaussian.

\begin{figure}[t!]
\centering
\includegraphics[width=0.48125\textwidth]{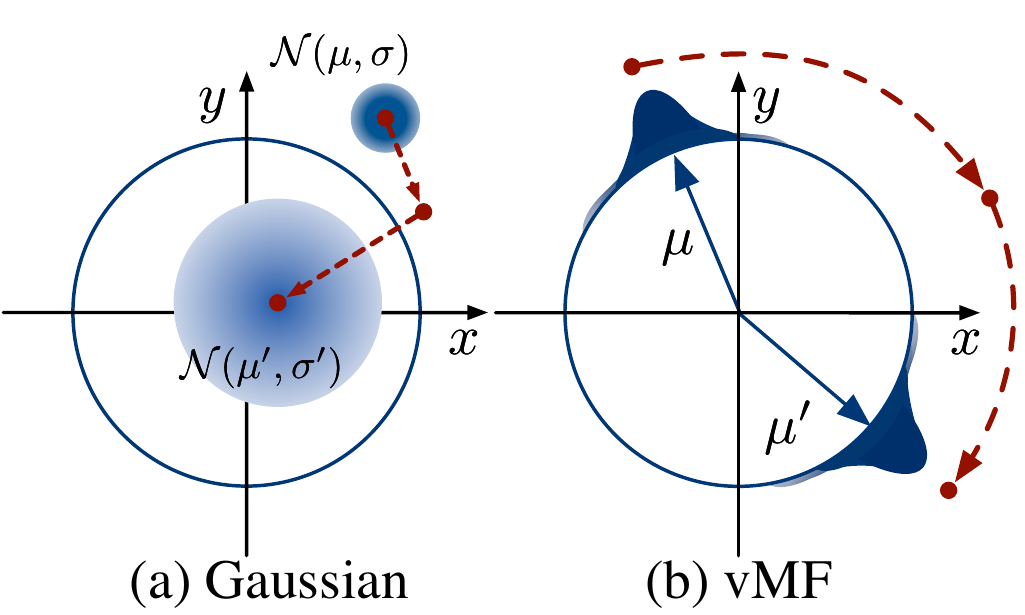}
\caption{Visualization of optimization of how $q$ varies over time for a single example during learning. In the Gaussian case, the KL term tends to pull the model towards the prior (moving from $\mu, \sigma$ to $\mu^{\ensuremath{\prime}}, \sigma^{\ensuremath{\prime}}$), whereas in the vMF case there is no such pressure towards a single distribution. 
} \label{klc}
\end{figure}

\paragraph{VAE with vMF} We will use vMF as both our prior and variational posterior in our VAE models. Otherwise, the setup for our VAE remains the same as in the Gaussian case established in Section~\ref{sec:bg}. Our prior is the uniform distribution vMF($\cdot,\kappa=0$). Since true posterior $p_{\theta}(z|x)$ is intractable, we will approximate it with a variational posterior $q_{\phi}(z|x) =  \text{vMF}(z; \mu, \kappa)$ where the mean direction $\mu$ is the output of encoding neural networks (Figure~\ref{model}, right side) and $\kappa$ is treated as a constant. 

Before we can implement a VAE, we need to derive an expression for KL divergence in order to optimize ELBO (Equation~\ref{elbo}) and give a sampling algorithm that admits the reparameterization trick \cite{kingma2013autoencoding}.

\paragraph{KL divergence} With vMF($\cdot,0$) as our prior, the KL divergence is:\footnote{Our KL divergence agrees with that of \newcite{Davidson_Falorsi_Cao_Kipf_Tomczak_2018} (see their appendix for a derivation), and we have verified it empirically. The equation in \newcite{guu2017generating} gives slightly different KL values, though differences are small ($<$5\%) for most $\kappa$ and dimension values we encounter.}
\begin{multline*}
\text{KL}(\text{vMF}(\mu,\kappa) || \text{vMF}(\cdot,0))  =\kappa \frac{ I_{d/2}(\kappa)}{I_{d/2-1}(\kappa)}\\
+ \left(\frac{d}{2}-1\right)\log\kappa - \frac{d}{2}\log(2\pi) - \log I_{d/2-1}(\kappa) \\ + \frac{d}{2} \log \pi + \log 2 - \log\Gamma\left(\frac{d}{2}\right)
\end{multline*}
Critically, this only depends on $\kappa$, not on $\mu$. $\kappa$ will be treated as a fixed hyperparameter, so this term will be constant for our model; KL collapse will therefore be rendered impossible.

Figure \ref{klc} shows a visualization of the learning trajectories of Gaussian and vMF VAE. For the Gaussian VAE, the KL divergence in the objective function tends to pull the posterior towards the prior centered at the origin and, therefore, make the optimization difficult as mentioned before. For the vMF VAE, given fixed $\kappa$, there is no such vacuous state and $\mu$ can vary freely.

\begin{figure}[t!]
\centering
\includegraphics[width=0.48125\textwidth]{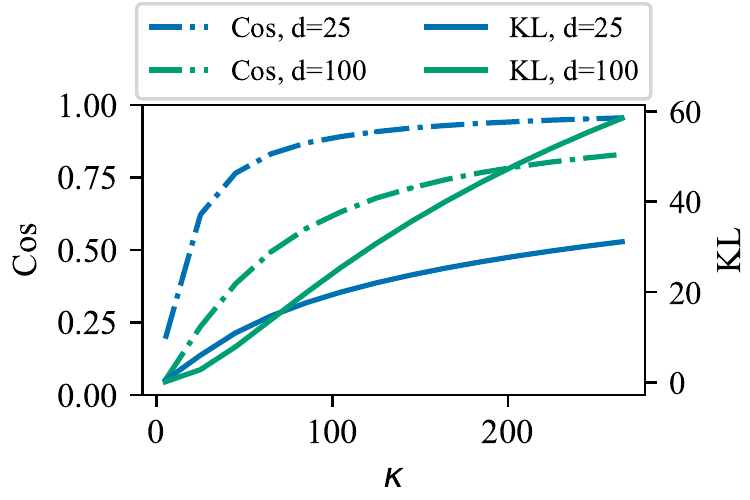}
\caption{Visualization of the interaction between $\kappa$, KL, and dimensionality in vMF. Cos represents the cosine similarity  between $\mu$ and samples from vMF$_d$($\mu,\kappa$) which reflects how disperse the distribution is. KL is defined as KL with a uniform vMF prior, KL(vMF$_d$($\mu,\kappa)||\text{vMF}(\cdot,0)$). Higher $\kappa$ values yield higher cosine similarities, but also higher KL costs.
} \label{stat_vmf}
\end{figure}

Figure \ref{stat_vmf} shows the KL value and concentration of vMF($\mu,\kappa$) for two different dimensionalities. KL increases monotonically with $\kappa$, as does concentration measured by cosine similarity. To get a fixed cosine dispersion as dimensionality increases, higher $\kappa$ values are needed, resulting in higher KL values.

\paragraph{Sampling from vMF} Following the implementation of \newcite{guu2017generating}, we use the rejection sampling scheme of \newcite{wood1994simulation} to sample a ``change magnitude'' $w$. Our sample is then given by $z = w \mu + v \sqrt{1-w^2}$, where $v$ is a randomly sampled unit vector tangent to the hypersphere at $\mu$. Neither $v$ nor $w$ depends on $\mu$, so we can now take gradients of $z$ with respect to $\mu$ as required.

\section{Experiments on Language Modeling}
\label{sec:nvrnn}

We first evaluate our vMF approach in the NVRNN setting. We will return to this model and analyze its properties further in Sections~\ref{ana:what} and \ref{ana:task} after showing experiments on document modeling.

\paragraph{Dataset}
For NVRNN, we use the Penn Treebank \cite{marcus1993building}, also used in \newcite{bowman2016generating}, and Yelp 2013 \cite{xu2016cached}. Examples in the Yelp dataset are much longer and more diverse than those from PTB, requiring more understanding of high-level semantics to generate a coherent sequence.
Yelp has a long tail of very long reviews, so we truncate the examples to a maximum length of 50 words; this still gives an average length over twice as long as in the PTB setting. Statistics about all datasets used in this paper are shown in Table \ref{dataset}.

\begin{table}[t!]
\begin{center}
\small
\begin{tabular}{ c|ccccc }
\toprule
Name & Train & Dev & Test & Len & Vocab\\
\midrule
PTB & 42068 & 3370 &  3761 & 21.1 & 10K \\ 
Yelp & 62522 & 7773 & 8671 & 49.5 & 15K \\ 
\midrule
20NG & 11268 & - & 7505 & 96.1 & 2K\\ 
RC & 794414 & - & 10000 &116.8 & 10K \\ 
\bottomrule
\end{tabular}
\end{center}
\caption{\label{dataset} Statistics of the datasets used in our experiments. Len stands for the average length of an example. Vocab is the vocabulary size; these follow prior work.}
\end{table}

\begin{table*}[t!]
  \centering
  \small
  \begin{tabular}{c|cccc|cccc}
  \toprule
  \multirow{3}{*}{Model} &
  \multicolumn{4}{c|}{PTB} &
  \multicolumn{4}{c}{Yelp}  \\
   &\multicolumn{2}{c}{Standard} & \multicolumn{2}{c|}{Inputless}  &\multicolumn{2}{c}{Standard} & \multicolumn{2}{c}{Inputless}\\
     & NLL & PPL & NLL & PPL & NLL & PPL & NLL & PPL \\
     \midrule
RNNLM \shortcite{bowman2016generating} & 100 (\ --\ ) & 116 & 135 (\ --\ ) & \textgreater 600 & -- & -- & -- & -- \\
G-VAE \shortcite{bowman2016generating} & 101 (2) & 119 & 125 (15) & 380 & -- & -- & -- & -- \\
\midrule
RNNLM (Ours) & 100 (\ --\ ) & 114 & 134 (\ --\ ) & 596  & 199 (\ --\ )  &  55  &  300 (\ --\ )  &  432\\
G-VAE (Ours) & \phantom{0}99 (4.4) & 109 & 125 (6.3) & 379 & 199 (0.5)  &  55  &  274 (13.4)  &  256 \\
vMF-VAE (Ours) & \phantom{0}\textbf{96	(5.7)}	&	\textbf{98} & \phantom{0}\textbf{117	(18.6)}	&	\textbf{262} & \textbf{198 (6.4)}  &  \textbf{54}  &  \textbf{242 (48.5)}  &  \textbf{134} \\
\bottomrule
  \end{tabular}
  \caption{Experimental results of NVRNN on the test sets of PTB and Yelp. The upper RNNLM and G-VAE shows the result from \newcite{bowman2016generating}. KL divergence is shown in the parenthesis, along with total NLL. Best results are in bold. vMF consistently uses higher KL term weights but achieves comparable or better NLL and perplexity values across all four settings.}
  \label{nvrnn}
\end{table*}

\paragraph{Settings}


We evaluate our NVRNN as in \newcite{bowman2016generating} and explore two different settings. In the \textbf{Standard} setting, the input to the RNN at each time step is the concatenation of the latent code $z$ and the ground truth word from the last time step, while the \textbf{Inputless} setting does not use the prior word. The more powerful decoder of the Standard setting makes the latent representations inherently less useful. In the Inputless setting, the decoder needs to predict the whole sequence with only the help of given latent code. In this case, a high-quality representation of the sentence is badly needed and the model is driven to learn it.

Our implementation of VAE uses a one layer unidirectional LSTM as both encoder and decoder. We use an embedding size of 100 and hidden units of size 400 in the LSTM. The dimension of the latent code is chosen from $\{ 25,50,100 \}$ by tuning on the development set. 
We use SGD to optimize all models with decayed learning rate and gradient clipping. For Yelp, the sentiment bit, which ranges from 1 to 5, is also embedded into a 50 dimension vector and input for every time step of the decoding phase.

\paragraph{Results}
Experimental results of the NVRNN are shown in Table~\ref{nvrnn}. We report negative log likelihood (NLL)\footnote{Reported values are actually a lower bound on the true NLL, computed from ELBO by sampling $z$.} and perplexity (PPL) on the test set.  We follow the implementation reported in \newcite{bowman2016generating} where the KL term weight is annealed for the Gaussian VAE; vMF VAE works well without weight annealing. The vMF distribution gives a performance boost in all datasets in both the Standard and Inputless settings. Even in the Standard setting, our model is able to successfully use nonzero KL values to achieve better perplexities, and even when KL collapse does not appear to be the case (e.g., G-VAE on the PTB-Standard setting), a Gaussian family of distributions results in lower KLs and worse log likelihoods, possibly due to optimization challenges. In the Inputless setting, we see large gains: vMF VAE reduces PPL from 379 to 262 in PTB, and from 256 to 134 in Yelp compared to Gaussian VAE.

\begin{figure}[t!]
\centering
\includegraphics[width=0.48125\textwidth]{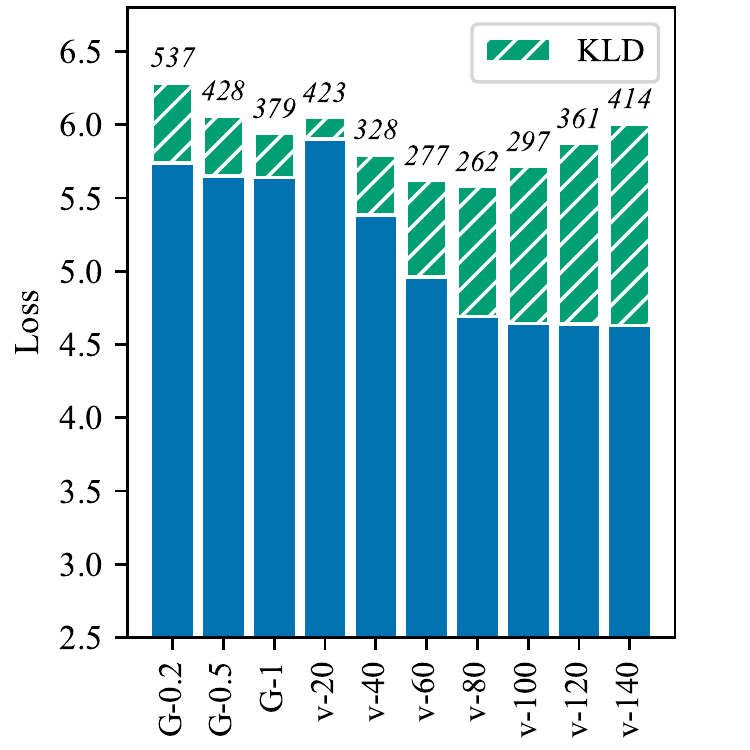}
\caption{Comparison of Gaussian- and vMF-NVRNN with different hyper-parameters. All models are trained on PTB in the Inputless setting where the latent dimension is 50. G-$\alpha$ indicates Gaussian VAE with KL annealed by the given constant $\alpha$, and V-$\kappa$ indicates VAE with $\kappa$ set to the given value. 
The green bar reflects the amount of KL loss while the total height reflects the whole objective.
\textit{Numbers} above bars are perplexity. vMF is more highly tunable and also achieves stronger results across a wide range of $\kappa$ values.} 
\label{kl-tradeoff}
\end{figure}

\paragraph{Trade-off Comparison}


Besides the overall perplexity, we are also interested in the trade-off between reconstruction loss and KL, and the contribution of KL to the whole objective. Figure~\ref{kl-tradeoff} shows the ability of our model to explicitly control the balance between the KL and the reconstruction term. First, we ``permanently'' anneal the Gaussian VAE by setting the weight of the KL term to a constant smaller than 1 (0.2 and 0.5 in our case). We find that this trick does mitigate the KL collapse, but the overall performance is worse. Therefore, this is not only a numerical game about the KL vs. NLL trade-off but a deeper challenge of how to structure models to learn effective latent representations.

For vMF VAE, when we gradually increase the value of $\kappa$, the concentration of the distribution around the mean direction $\mu$ is higher and samples from vMF are closer to $\mu$. The model achieves the best perplexity when $\kappa=80$. The reconstruction error is bounded around 4.5 due to the difficulty of the task and limited capacity of LSTM decoder. While $\kappa$ is a hyperparameter that needs to be tuned, the model is overall not very sensitive to it, and we show in Section~\ref{ana:task} that reasonable $\kappa$ values transfer across similar tasks.

\section{Experiments on Document Modeling}

We also investigate how vMF VAE performs in a different setting, one less plagued by the KL collapse issue. Specifically, the Neural Variational Document Model (NVDM), proposed by \newcite{miao2016neural}, is a VAE-based unsupervised document model. This model follows the VAE framework introduced in Section~\ref{sec:bg}. Our document representation is an indicator vector $x$ of word presence or absence in the document. Since this is a fixed-size representation, we use 2-layer MLPs with 400 hidden units for both the encoder $q(z|x)$ and decoder $p(x|z)$; the decoder places a simple multinomial distribution over words in the vocabulary, and the probability of a document is the product of the probabilities of its words.

\begin{table}[t]
\begin{center}
\small
\begin{tabular}{ c|c|c|c }
\toprule
Model & Dim & 20NG & RCV1\\
\midrule
\multirow{2}{*}{fDARN \shortcite{mnih2014neural}} & 50 & 917 & 724 \\
 & 200 & - & 598 \\\midrule
\multirow{2}{*}{G-NVDM \shortcite{miao2016neural}} & 50 & 836 & 563 \\
 & 200 & 852& 550\\ \midrule
\multirow{3}{*}{v-NVDM (Ours)} & 25 & \textbf{793} &  558\\
 & 50 & 830 & \textbf{529} \\
 & 200 & 851 & 609 \\ 
\bottomrule
\end{tabular}
\end{center}
\caption{\label{nvdm} Test set perplexities for the document modeling task. Feedforward Deep Auto Regressive Neural Network (fDARN) is implemented by \newcite{mnih2014neural}. Gaussian-based NVDM (G-NVDM) is proposed in \newcite{miao2016neural}. Dim indicates the dimension of the latent code. Our v-NVDM model outperforms past models by a substantial margin.}
\end{table}

\paragraph{Dataset}

For NVDM, we use two standard news corpus, 20 News Groups (20NG) and the Reuters RCV1-v2, which were used in \newcite{miao2016neural}.\footnote{The preprocessed version can be downloaded from \href{https://github.com/ysmiao/nvdm}{\texttt{https://github.com/ysmiao/nvdm}}}


\paragraph{Results}
Experimental results\footnote{We do not compare to results from \newcite{serban2017piecewise}. Compared to our current results, that work reports very strong performance on 20NG and very weak performance on RCV1. Based on consultation with the authors, they use different preprocessing than \newcite{miao2016neural}.} are shown in Table~\ref{nvdm}. In contrast with NVRNN, the NVDM fully relies on the power of latent code to predict the word distribution, so we never observe a KL collapse, yet vMF still achieves better performance than Gaussian. As shown in Figure \ref{stat_vmf}, in order to keep the same amount of dispersion in samples from the variational posterior, larger latent dimensions need larger $\kappa$ values and correspondingly larger KL term values. For 20NG, which is much smaller than RCV1, smaller dimensions therefore give better performance. For both datasets, the settings of $\kappa=100,\text{dim}=25$ and $\kappa=150,\text{dim}\in\{50,200\}$ work well.

\begin{table}[t]
  \centering
  \small
  \begin{tabular}{c|cccc}
  \toprule
  \multirow{2}{*}{$P(x |z, \text{BoW})$} 
   &\multicolumn{2}{c}{Standard} & \multicolumn{2}{c}{Inputless}  \\
     & NLL & PPL & NLL & PPL  \\
\midrule
RNNLM  & 79 (--)  &  43 &  106 (--)  &  152  \\
G-VAE  & 79 (0.0)  &  43	&  106 (0.4)  &  153  \\
v-VAE  &  \textbf{73	(0.2)}	& \textbf{33} &  \textbf{93	(11.4)}	&	\textbf{82} \\
\bottomrule
  \end{tabular}
  \caption{Experimental results of NVRNN-BoW on PTB; i.e., the decoder also conditions on a bag of words representation of the sentence to generate. In this case, the Gaussian models exhibit KL collapse but vMF can still learn effectively.}
  \label{bow}
\end{table}

\section{What do our VAEs encode?}
\label{ana:what}
We design more probing tasks to demonstrate what is encoded in latent representations induced by vMF VAE. One additional model variant we explore here is the NVRNN-BoW model. This is a variant of NVRNN where the decoder additionally conditions on the vector $BoW = \frac{1}{n} \sum_{i=1}^n e(x_i)$, the average word embedding value of the sentence $x$. While an artificial setting, this lets us see how effectively the latent code can capture information other than simple word choice by making a form of this information independently available. Table~\ref{bow} shows results in this setting, where once again we see the KL collapse problem for the Gaussian models and better performance from vMF on perplexity in both the Standard and Inputless settings.


\paragraph{Is the latent code more than a bag of words?} For all of these models, one hypothesis is that the encoder may be learning to memorize the bag of words and then preferentially generate words in that bag from the decoder. To verify this, we investigate whether the BoW representation and the learned latent code can be reconstructed from each other. Specifically, given a sentence $x$ we can compute $BoW$ as defined above and $\mu = \textrm{enc}(x)$, the latent encoding of $x$ as represented by the mean vector output by the encoder. We can use a simple multilayer perceptron to to try to map from the bag of words to the latent code: $\hat{\mu} = MLP(BoW)$, then learn the parameters of the MLP by minimizing $\Vert \hat{\mu} - \mu \Vert^2$ on a sample. The same process can be used to learn a mapping from $\mu$ back to the bag of words.

Table~\ref{c2b} shows averaged cosine similarities of our reconstructions under both Gaussian and vMF models. For vMF, $\mu$ can reconstruct the bag-of-words more accurately than the bag-of-words can reconstruct $\mu$, indicating that the latent code in vMF captures more information beyond the bag of words.

\begin{table}[t]
\begin{center}
\small
\begin{tabular}{ c|cc|c }
\toprule
Model & \multicolumn{2}{c|}{NVRNN} & NVRNN-BoW \\
Setting & $\mu\rightarrow$ BoW & BoW $\rightarrow\mu$  & $\mu\rightarrow$ BoW  \\
\midrule
G-VAE & 0.74 & 0.74 &  0.32  \\ 
v-VAE & 0.77 &  0.57 & 0.23  \\ 
\bottomrule
\end{tabular}
\end{center}
\caption{Average cosine similarity when trying to reconstruct the latent code $\mu$ from the bag of words and vice versa. In vMF, the latent code contains more information beyond the bag of words, as shown by the lower cosine similarity when predicting BoW $\rightarrow\mu$ (0.57). When the latent code is learned in a model conditioned on the bag of words (right column), it predicts the bag of words much less well, indicating that the model successfully learns orthogonal information.}
\label{c2b} 
\end{table}

We repeat this experiment in a separate NVRNN model where the decoder can explicitly condition on the $BoW$ vector described above. The results are shown in the right column of Table~\ref{c2b}. Our model, v-VAE, achieves a lower cosine similarity than G-VAE (0.23 vs.~0.32), indicating that it capturing less redundant information and using the latent space to more efficiently model other properties of the data.

\paragraph{Sensitivity to word order} Table~\ref{c2b} shows that NVRNN with vMF encodes information beyond the bag of words; a natural hypothesis is that it is encoding word order. We can more directly investigate this in the context of both NVRNN and NVRNN-BoW settings. Inspired by \newcite{junbo2017adversarially}, we propose an experiment probing the sensitivity to randomly swapping adjacent pairs of words for the encoding in the Inputless setting on PTB. We vary the probability of swapping each word pair and see how the latent code changes as the number of swaps increases. Ideally, our models should capture ordering information and therefore be sensitive to this change.

\begin{figure}[t]
\centering
\includegraphics[width=0.48125\textwidth]{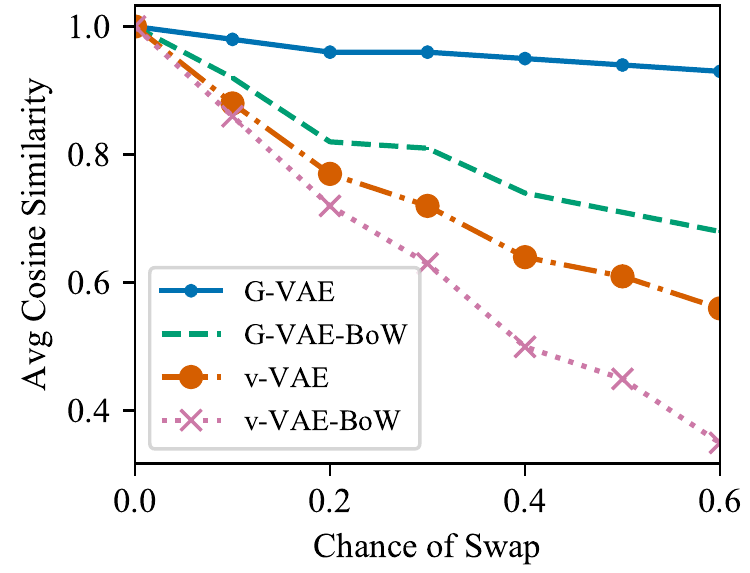}
\caption{Sensitivity of latent codes to swapping adjacent words of encoding sequence. Cosine similarity is measured between the latent code (encoded mean vector) of the original sentence and the sentence after swaps are applied. We see that vMF is more highly sensitive to swaps in both the NVRNN and NVRNN-BoW settings, indicating that its latent space likely encodes more ordering information.} \label{word-order}
\end{figure}

Figure \ref{word-order} shows the results. v-VAE's representations are more sensitive than those of the G-VAE: they change faster as swaps become more likely.\footnote{The Gaussian VAE here makes very little use of the latent variable, hence why the representations change very little.} In the NVRNN-BoW setting, we see that the models are even more sensitive. vMF enables us to more easily learn this kind of desirable information in our sentence encodings.

\section{Controlling Variance with $\kappa$}
\label{ana:task}

A core aspect of our approach so far has been treating $\kappa$ as a fixed hyperparameter. Fixing $\kappa$ is beneficial from an optimization standpoint: it makes it more difficult for the model to get stuck in local optima. But it also reduces the model's flexibility, since we can no longer predict per-example $\kappa$ values, and it introduces another parameter that the system designer must tune.

\begin{figure}[t]
\centering
\includegraphics[width=0.48125\textwidth]{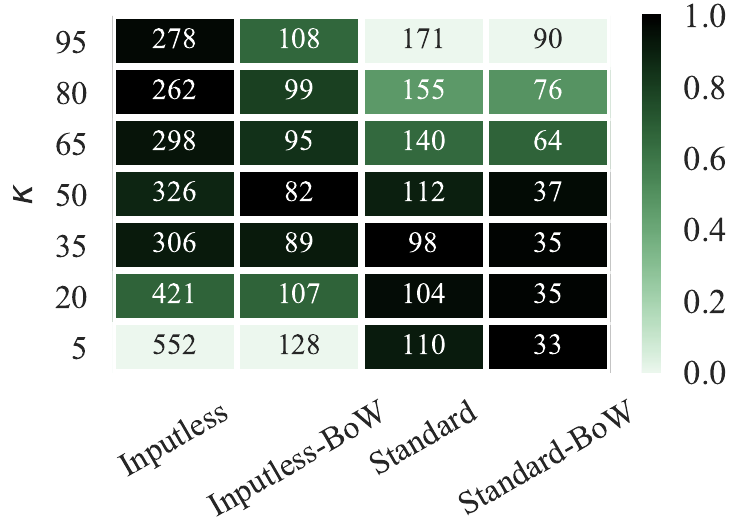}
\caption{Perplexity of v-VAE in different settings with different $\kappa$ values when the latent dimension is 50. Darker colors correspond to perplexity values closer to the best observed for that setting. For each task, we see that there is a range of $\kappa$ values that work well, and these transfer between comparable tasks.}
\label{heatmap}
\end{figure}

Fortunately, a wide range of $\kappa$ values appear to work well for the tasks we consider. Figure~\ref{heatmap} shows how the concentration parameter $\kappa$ changes the results on PTB when the latent dimension and other hyperparameters are held fixed. We have ordered the tasks left-to-right from ``hardest'' to ``easiest'' in terms of necessity of latent representation: the Inputless setting needs heavy information from the latent code to reconstruct the sentence, whereas the Standard-BoW setting has an extremely strong decoder to predict the next word. We see that in each of these cases, a wide range of $\kappa$ values works, and moreover reasonable $\kappa$ values transfer between the two Standard and between the two Inputless settings, indicating that the overall approach is not highly sensitive to these hyperparameter values.

\paragraph{Brittleness of Learning $\kappa$} Throughout this work, we have treated $\kappa$ as a fixed parameter. However, we can treat $\kappa$ in the same way as $\sigma$ in the Gaussian case and learn it on a per-instance basis. The KL divergence of vMF is differentiable with respect to $\kappa$ given gradients of the modified Bessel function of the first kind,\footnote{$\nabla_{\kappa} I_{d}(\kappa) = \frac{1}{2}( I_{d-1}(\kappa)  + I_{d+1}(\kappa))$} allowing us to change the concentration on a per-instance basis. However, this reintroduces the issue of KL collapse: the KL term will encourage $\kappa$ to be as low as possible, potentially making the latent variable vacuous.

In practice, we observe that it is necessary to clip $\kappa$ values to a certain range for numerical reasons. Within this range, the model gravitates towards the smallest $\kappa$ values and performs substantially worse than models trained with our fixed $\kappa$ approach. This indicates that even with the vMF model, the optimization problem posed by ELBO is simply a hard one and the approach of fixing KL divergence is a surprisingly good optimization technique.

\section{Related Work}

\paragraph{Applications of VAE in NLP} 
Deep generative models have achieved impressive successes in domains adjacent to NLP such as image generation \cite{gregor2015draw,oord2016pixel} and speech generation \cite{chung2015a,oord2016wavenet}. VAEs specifically \cite{kingma2013autoencoding,rezende2014stochastic} have been a popular model variant in NLP. They have been applied to tasks including document modeling \cite{miao2016neural}, language modeling \cite{bowman2016generating}, and dialogue generation \cite{serban2017a}. VAEs can be also be applied for semi-supervised classification \cite{xu2017variational}. Recent twists on the standard VAE approach including combining VAE and holistic attribute discriminators for conditional generation \cite{hu2017toward} and using a more flexible latent space regularized by an adversarial method \cite{junbo2017adversarially}. 

\paragraph{VAE Objective} Several pieces of recent work have highlighted the issues with optimizing the VAE objective. \newcite{Alemi_Poole_Fischer_Dillon_Saurous_Murphy_2018} shed light on the problem from the perspective of information theory. \newcite{Zhao_Song_Ermon_2017} and \newcite{betavae} both propose various reweightings of the objective along with theoretical and empirical justification.

\paragraph{Choices of Priors for VAE} Some past work has explored various priors for VAE. \newcite{serban2017piecewise} proposed a piecewise constant distribution which deals with multiple modes, but which sacrifices the property of continuous interpolation. \newcite{guu2017generating} also applied vMF in a VAE model, but used theirs specifically in the sentence-editing case. \newcite{Davidson_Falorsi_Cao_Kipf_Tomczak_2018} explored vMF in a VAE model for MNIST and a link prediction task.  \newcite{Hasnat_Bohné_Milgram_Gentric_Chen_2017} applied the vMF distribution for facial recognition. 
Other past work has used different decoders, including CNNs \cite{yang2017improved} and CNN-RNN hybrids \cite{semeniuta2017a}. Changing the decoder is a change largely orthogonal to changing the prior: it can alleviate the KL vanishing issue, but it does not necessarily scale to new settings and does not give explicit control over utilization of the latent code.

\section{Conclusion}
In this paper, we propose the use of a von Mises-Fisher VAE to resolve optimization issues in variational autoencoders for text. This choice of distribution allows us to explicitly control the balance between the capacity of the decoder and the utilization of the latent representation in a principled way. Experimental results demonstrate that the proposed model has better performance than a Gaussian VAE across a range of settings. Further analysis shows that vMF VAE is more sensitive to word order information and makes more effective use of the latent code space.

\section*{Acknowledgments}
This work was partially supported by NSF Grant IIS-1814522, a Bloomberg Data Science Grant, and an equipment grant from NVIDIA. The authors acknowledge the Texas Advanced Computing Center (TACC) at The University of Texas at Austin for providing HPC resources used to conduct this research.  Thanks as well to the anonymous reviewers for their helpful comments.

\bibliography{ms}
\bibliographystyle{acl_natbib_nourl}

\end{document}